\documentclass[acmtog]{acmart}
\acmSubmissionID{1016}

\usepackage{booktabs} 
\usepackage{soul}
\usepackage{enumitem}
\usepackage{nicefrac}
\usepackage{tikz}

\newcommand{\modelName}{\textit{AutoBrep}}

\usepackage{xcolor}
\usepackage{amsmath}

\usepackage{graphicx}
\usepackage{subcaption}
\usepackage{mwe}

\usepackage[ruled]{algorithm2e} 

\SetAlFnt{\small}
\SetAlCapFnt{\small}
\SetAlCapNameFnt{\small}
\SetAlCapHSkip{0pt}

\acmJournal{TOG}

\copyrightyear{2025}
\acmYear{2025}
\setcopyright{acmlicensed}\acmConference[SA Conference Papers '25]{SIGGRAPH Asia 2025 Conference Papers}{December 15--18, 2025}{Hong Kong, Hong Kong}
\acmBooktitle{SIGGRAPH Asia 2025 Conference Papers (SA Conference Papers '25), December 15--18, 2025, Hong Kong, Hong Kong}
\acmDOI{10.1145/3757377.3763814}
\acmISBN{979-8-4007-2137-3/2025/12}

\begin{document}

\title{AutoBrep: Autoregressive B-Rep Generation with Unified Topology and Geometry}

\author{Xiang Xu}
\orcid{0000-0002-3437-1470}
\affiliation{%
 \institution{Autodesk Research}
 \city{Toronto}
 \country{Canada}}
\email{xiang.xu@autodesk.com}
\authornote{Equal contribution}

\author{Pradeep Kumar Jayaraman}
\orcid{0000-0001-6314-6136}
\affiliation{%
 \institution{Autodesk Research}
 \city{Toronto}
 \country{Canada}}
\email{pradeep.kumar.jayaraman@autodesk.com}
\authornotemark[1]

\author{Joseph G. Lambourne}
\orcid{0000-0002-9892-1945}
\affiliation{%
 \institution{Autodesk Research}
 \city{London}
 \country{UK}}
\email{joseph.lambourne@autodesk.com}
\authornotemark[1]

\author{Yilin Liu}
\orcid{0000-0001-7336-1956}
\affiliation{%
 \institution{Autodesk Research}
 \city{London}
 \country{UK}}
\email{yilin.liu@autodesk.com}

\author{Durvesh Malpure}
\orcid{0009-0001-7326-6526}
\affiliation{%
 \institution{Autodesk Research}
 \city{San Francisco}
 \country{USA}}
\email{durvesh.malpure@autodesk.com}

\author{Pete Meltzer}
\orcid{0000-0003-2496-5117}
\affiliation{%
 \institution{Autodesk Research}
 \city{London}
 \country{UK}}
\email{pete.meltzer@autodesk.com}

\renewcommand\shortauthors{X. Xu, P. Jayaraman, J. Lambourne, Y. Liu, D. Malpure, and P. Meltzer}

\begin{abstract}
The boundary representation (B-Rep) is the standard data structure used in Computer-Aided Design (CAD) for defining solid models. Despite recent progress, directly generating B-Reps end-to-end with precise geometry and watertight topology remains a challenge. This paper presents \modelName, a novel Transformer model that autoregressively generates B-Reps with high quality and validity. \modelName\ employs a unified tokenization scheme that encodes both geometric and topological characteristics of a B-Rep model as a sequence of discrete tokens. Geometric primitives (i.e., surfaces and curves) are encoded as latent geometry tokens, and their structural relationships are defined as special topological reference tokens. Sequence order in \modelName\ naturally follows a breadth first traversal of the B-Rep face adjacency graph. At inference time, neighboring faces and edges along with their topological structure are progressively generated. Extensive experiments demonstrate the advantages of our unified representation when coupled with next-token prediction for B-Rep generation.  \modelName\ outperforms baselines with better quality and watertightness. It is also highly scalable to complex solids with good fidelity and inference speed. We further show that autocompleting B-Reps is natively supported through our unified tokenization, enabling user-controllable CAD generation with minimal changes. Code is available at \url{https://github.com/AutodeskAILab/AutoBrep}.

\end{abstract}

%
%


%
%


\begin{teaserfigure}
 \centering
\includegraphics[width=0.98\textwidth]{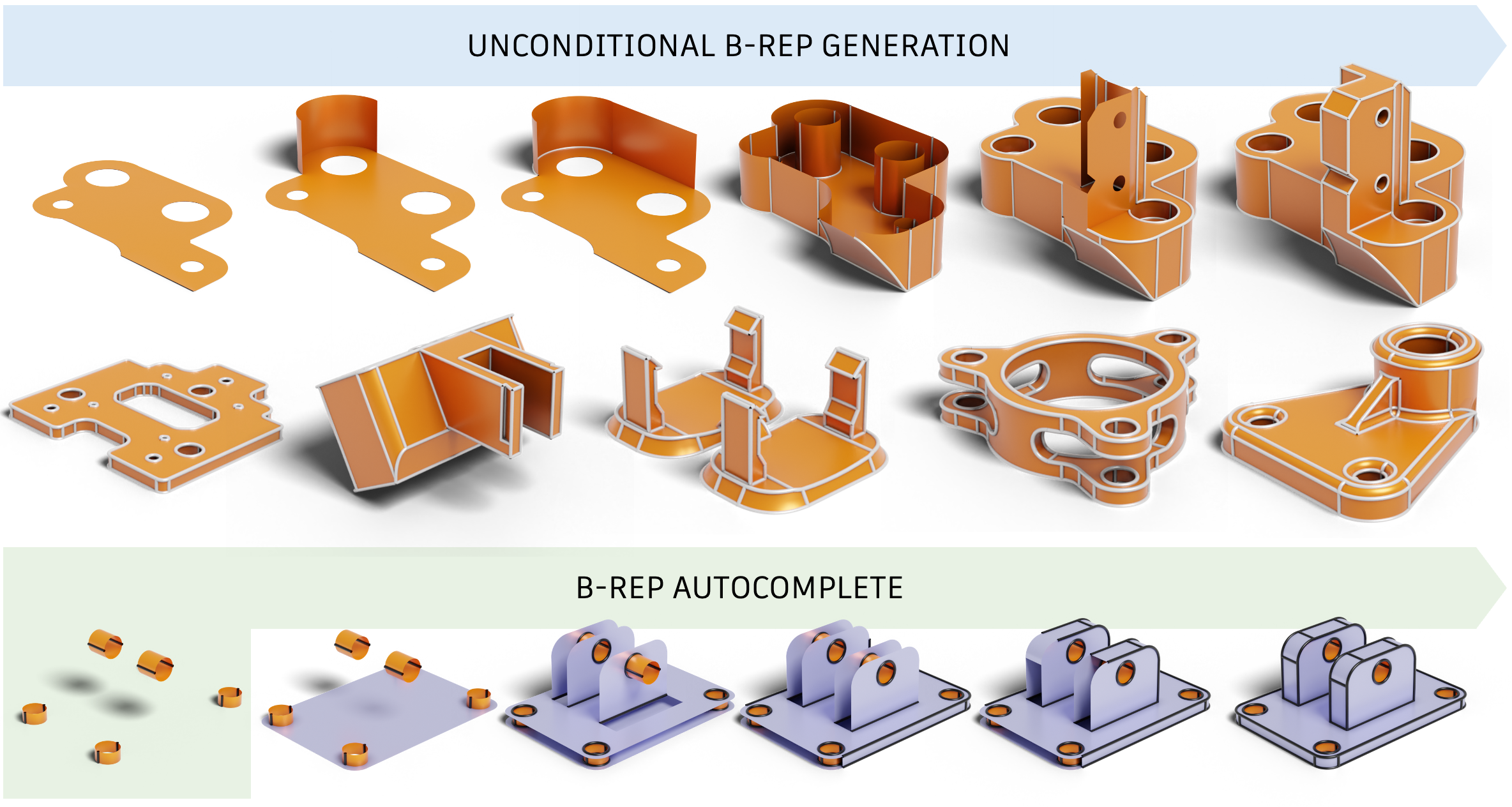}
  \caption{\modelName\ is a single autoregressive Transformer that progressively generates B-Rep geometry and topology following a breadth-first traversal of the face adjacency graph. The model inherently supports B-Rep autocompletion, guaranteeing that the output geometry faithfully preserves the conditional faces.}
  \label{fig:teaser}
\end{teaserfigure}

\let\subsectionautorefname\sectionautorefname
\let\subsubsectionautorefname\sectionautorefname

\maketitle

\newcommand{\predF}{\tilde{F}}
\newcommand{\Fxyz}{F^\text{xyz}}
\newcommand{\predFxyz}{\tilde{F}^\text{xyz}}
\newcommand{\Fnor}{F^\text{nor}}
\newcommand{\predFnor}{\tilde{F}^\text{nor}}
\newcommand{\umin}{u_\text{min}}
\newcommand{\umax}{u_\text{max}}
\newcommand{\vmin}{v_\text{min}}
\newcommand{\vmax}{v_\text{max}}
\newcommand{\dFdu}{\partial \predFxyz_u}
\newcommand{\dFdv}{\partial \predFxyz_v}
\newcommand{\Lxyz}{\mathcal{L}_\text{xyz}}
\newcommand{\Lnor}{\mathcal{L}_\text{nor}}
\newcommand{\Lcons}{\mathcal{L}_\text{cons}}
\newcommand{\cosine}{\text{Cosine}}

\section{Introduction}
\label{sec:intro}
Computer-Aided Design (CAD) plays a crucial role in the creation of engineered objects that are part of our daily lives.
A solid model in CAD is commonly represented as a Boundary Representation (B-Rep), an interlinked graph-like structure comprising topological vertices, edges, loops, faces, that contain geometric points, curves and surfaces~\cite{weiler1986}. This representation is vital for constructing precise 3D models suitable for simulation and manufacturing.


The ability to automatically create high-quality B-Reps has the potential to revolutionize traditional CAD design workflows. Recent advances in deep learning have enabled two orthogonal approaches for B-Rep generation.
Sequential generation~\cite{wu2021deepcad,xu2022skexgen,xu2023hierarchical,li2025cad} constructs a sequence of parametric modeling features, i.e., sketches, extrusions and revolutions, that can be rebuilt into B-Reps using solid modeling kernels. This approach is advantageous for ensuring watertightness and editability, but is constrained by the need for modeling sequence datasets, which are typically limited in size. Significant effort is also required to support additional features such as fillets and chamfers.
Direct  generation~\cite{jayaraman2022solidgen,xu2024brepgen,fan2024neuronurbs,HolaBRep25,DTGBrepGen} bypasses this data limitation by working with raw B-Reps and directly synthesizing entities such as faces and edges that can be sewn into a solid model. 
This approach naturally supports complex geometry including freeform surfaces and benefits from availability of large-scale data. However, guaranteeing watertightness and validity of results are challenging as the predicted B-Rep elements must align precisely with one another. 

Direct generation methods have historically used multi-stage training. SolidGen~\cite{jayaraman2022solidgen} uses three separate models to generate the vertices, edges, and faces. Each model is trained independently with teacher forcing~\cite{williams89-teacherforcing} and cascaded at inference.
BrepGen~\cite{xu2024brepgen} has six modules: two autoencoders for the faces and edges, and four diffusion models to generate the geometry. Recent concurrent works~\cite{DTGBrepGen,wu2025cmtcascademartopology} have followed suit. The use of multiple stages is necessary due to the complexities involved in generating heterogeneous geometries (such as surfaces, curves and bounding boxes) and the topology required to connect them, which is inherently discrete and combinatorial.
Additionally, it is difficult to scale up multi-stage approaches due to complex dependencies across modules.
Cascading separate modules at inference increases error accumulation and limits the model's ability to learn the joint distribution of B-Rep entities, leaving out much of the underlying priors.

In this paper, we propose \modelName, an autoregressive Transformer model that represents the faces, edges, and topology of B-Reps as a unified sequence of discrete tokens. 
Building on the work of~\cite{xu2024brepgen},  we extract B-Rep faces and edges as point grids uniformly sampled in the parameter domain of their underlying surfaces and curves.
A key innovation in \modelName~ is the use of discrete representation learning to encode these point grids into compact latent tokens that are invariant to the parametrization. 
We also propose a novel topology tokenization that encodes the B-Rep face-edge incidence relationships to a set of local reference tokens.
Our proposed tokenization allows unifying geometry and topology into a single stream of tokens that can be  jointly modeled via next-token prediction (see \autoref{fig:teaser}). Experiments demonstrate that \modelName\ increases generation quality and reduces inference time. The use of a GPT-like Transformer model and discrete tokens also makes it naturally scalable to longer B-Rep sequences, allowing the generation of solids with unprecedented complexity. Finally, we demonstrate that B-Rep autocompletion is a special case in our unified tokenization scheme, where we can guarantee that the user faces are preserved in the generated valid B-Reps. 
In summary, our main contributions and improvements over state-of-the-art are:
\begin{itemize}[leftmargin=.25in]
  \item A tokenization scheme that encodes both the B-Rep geometry and topology as a single sequence of unified discrete tokens. 
  \item The use of a single autoregressive Transformer trained with next-token prediction for direct B-Rep generation, achieving higher quality than multi-stage diffusion, while having a 2$\times$ to 5$\times$ inference speedup and simplified training pipeline. 
  \item Scalability to complex B-Reps compared to just ~50 faces in the baselines.  This increased complexity is achieved while maintaining a higher fraction of valid and watertight B-Reps.
  \item Natural applicability to B-Rep autocompletion with guaranteed preservation of user provided faces in the results.
\end{itemize}

\section{Related Work}
\label{sec:related_work}

This section reviews CAD generation methods focusing on sketch-and-extrude and direct generation methods. A comprehensive review can be found in~\cite{Zhou2025review}.

\subsection{Sketch-Extrude and Constructive Solid Geometry}
\label{sec:sketch_extrude}
In traditional CAD workflows, designers construct 3D models by first creating 2D sketches and extending them into 3D using operations like extrusions and revolutions. With the availability of public CAD datasets~\cite{koch2019:abc,wu2021deepcad,willis2020fusion}, earlier learning-based methods like DeepCAD~\cite{wu2021deepcad} followed the sketch-and-extrude paradigm and use Transformer-based architecture to generate a sequence of sketch and extrude commands. 
Subsequent methods like SkexGen~\cite{xu2022skexgen}, HNC~\cite{xu2023hierarchical}, and CADTrans~\cite{guo2025cadtransn} introduce hierarchical codebooks to better capture local and global features.
This has also been extended to conditional generation tasks such as text-to-CAD~\cite{khan2024textcad}, image-to-CAD~\cite{you2024img2cad,chen2024img2cad,alam2024gencad}, voxel-to-CAD~\cite{voxel2cad22}, and multimodal generation~\cite{xu2024cadmllm}. Sketch2CAD~\cite{li2020sketch2cad} integrates sketch-based inputs into the construction sequence, while CADEditor~\cite{yuan2025cadeditor} and FlexCAD~\cite{flexcad} align textual operations with CAD editing tasks.
Despite their success, sketch-and-extrude methods are inherently constrained by the finite set of supported operations and primitives. 
Furthermore, they rely on construction history data, which is often unavailable. The DeepCAD dataset~\cite{wu2021deepcad}, which includes construction sequences, is five times smaller than the ABC dataset~\cite{koch2019:abc}, which only provides the final B-Rep models. 
Limitation in data size and operational flexibility hinders its scalability to larger-scale CAD generation.
Another direction is Constructive solid geometry (CSG)~\cite{DuIPSSRSM18,TianLSEFTW19,SharmaGLKM18,EllisNPSTS19,KaniaZK20,bspnet20,capri22,d2csg23}, which models shapes as combinations of simpler primitives like spheres, cubes, and cylinders. Despite usage in CAD reconstruction, CSG methods introduce ambiguities due to non-unique CSG trees for a given shape, which complicates generative learning~\cite{nvd24}.

\subsection{Direct Generation}
\label{sec:direct_generation}
Direct generation methods have gained popularity as they bypass the need for construction history and predefined operations, enabling the synthesis of arbitrary B-Rep models. 
SolidGen~\cite{jayaraman2022solidgen} first achieved this by decomposing B-Rep generation into vertex, edge, and face prediction, using a pointer network to model dependencies. 
However, the lack of support for freeform surface and curve types significantly limits its generalizability.
Building on UV-Net~\cite{jayaraman2021uvnet}, which introduces a uniform grid representation for faces and edges, BrepGen~\cite{xu2024brepgen} employs a multi-stage diffusion model for latent B-Rep generation. 
NeuroNURBS~\cite{fan2024neuronurbs} replaces UV grids with B-spline surface primitives for compact geometric encoding. 
DTGBrepGen~\cite{DTGBrepGen} further separates topology and geometry generation.
Despite advancements in direct generation methods, their performance remains constrained by the inherently multi-stage nature of B-Rep generation, where faces, edges, and vertices are tightly interdependent. 
The parametric structure of B-Reps is also highly sensitive to noise and misalignment which are exacerbated in multi-stage pipelines due to cumulative error. 
Moreover, the strong coupling between modules hinders scalability, 
as downstream components often require retraining when upstream stages are modified. 
To address this problem, concurrent work HoLa~\cite{HolaBRep25} introduces a holistic VAE that encodes entire B-Rep models into a unified latent space, simplifying the training process for the latent diffusion model.
In contrast, our method reduces the number of modules from six to three (two for encoding) and uses an autoregressive model to generate B-Reps without the need for padding, leading to substantial improvements in both generation quality and efficiency.



\begin{figure}
    \centering
    \includegraphics[width=0.94\columnwidth]{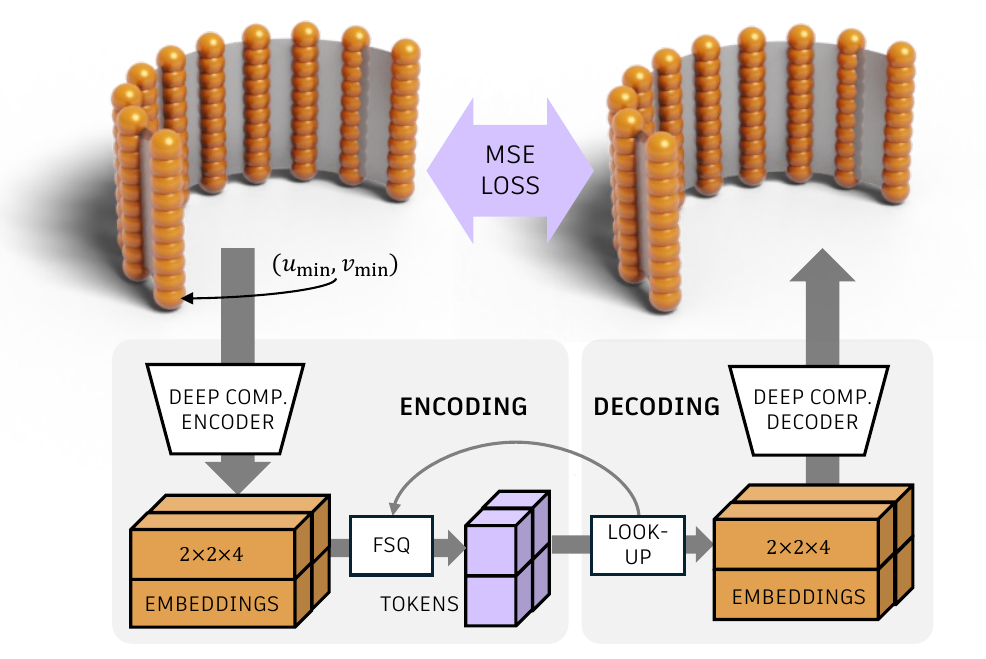}
    \caption{Geometric representation learning. Each face is represented by a 32$\times$32 point grid sampled uniformly in the parameter domain of the underlying surface, while ensuring the $(\umin,\vmin)$ point is the lowest lexicographically. Deep compression~\cite{dcae} is used to compress the point grid into low dimensional latent embeddings that are discretized using finite scalar quantization~\cite{fsq} into integer tokens.}
    \label{fig:geometry-learning}
\end{figure}

\begin{figure*}[ht]
    \begin{center}
    \includegraphics[width=0.9\textwidth]{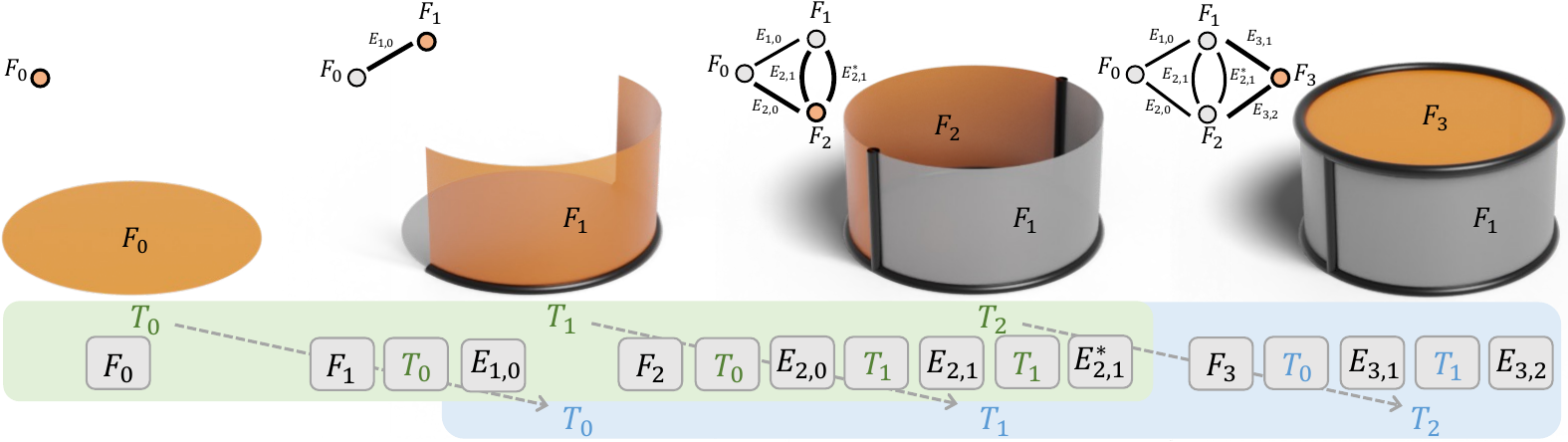}
        \caption{Top row: Sequence ordering follows a breadth-first traversal (BFT) of the B-Rep face adjacency graph (top-left). A new face (orange) and the connecting edge(s) are added incrementally during traversal. Previously visited faces are colored in gray. Bottom row: Unified discrete tokens include face geometry $F$, edge geometry $E$, and topology reference token $T$ for the face-edge incidence. $T$ is dynamically assigned to every face within a local context window for unique face references. This local window corresponds to the BFT level and is updated when traversal proceeds to the next window (from green to blue). * indicates the two vertical edges connecting the same set of semi-cylindrical faces.}
        \label{fig:data}
    \end{center}
\end{figure*}

\section{Geometry Representation Learning} 
\label{sec:geom-representation}
\subsection{Point Grids}
Following BrepGen~\cite{xu2024brepgen}, we adopt point grids sampled uniformly in the parameter domain of the underlying surface (resp. curve) in each B-Rep face (resp. edge) as the core geometric representation.
This representation provides a unified input feature for all the different parametric surfaces (resp. curves) without having to deal with specific parameters and equations.
Points are sampled on a 2D $N\times N$ equally spaced grid, denoted as $\Fxyz$, in the face's parameter domain $[\umin, \umax]\times[\vmin, \vmax]$.
Similarly, $N$  points are sampled on a 1D grid for edges on the $[\umin, \umax]$ parameter domain, with $N=32$ in all our experiments.

\subsection{Discrete Representations with FSQ}
\label{sec:fsq}
The regular structure of the point grids motivated BrepGen~\cite{xu2024brepgen} to utilize Stable Diffusion's KL-regularized variational autoencoder (VAE)~\cite{rombach2022StableDiffusion} for compressing the point grids into continuous grids of latent embeddings.
Each $(32 \times 32 \times 3)$ face point grid is compressed into a $(4 \times 4 \times 3)$ grid using 2D convolutional layers, while each $(32 \times 3)$ edge point grid is compressed into a $(4 \times 3)$ grid using 1D convolutional layers.
In this paper, we introduce two modifications to the representation learning.

Firstly, while BrepGen produces $(4\times 4 \times 3)$ embeddings per face, we aim to further compress this, as the resolution of the latent grid directly influences the number of tokens to be generated in our prior model.
For this, we replace the Stable-Diffusion encoder and decoder modules with the Deep Compression Autoencoder~\cite{dcae} which has been highly effective with compressing images without loss of reconstruction accuracy~\cite{sana}.
This allows us to further compress the face point grids to $2\times 2 \times 4$ latent grids, and edge points grids to $2\times 4$ grids.

Secondly, we employ finite-scalar quantization (FSQ)~\cite{fsq} (and remove the KL regularization) to learn discrete codes from the latent grid of continuous embeddings produced by the autoencoder to facilitate autoregressive generation.
FSQ is an alternative to vector quantization~\cite{vqvae} with the benefit that codebook usage is automatically maximized without the need for auxiliary losses.
FSQ performs rounding of each continuous value along each dimension of the continuous embeddings, based on a predetermined set of levels.
In our case, we use $L=[8, 5, 5, 5]$ levels yielding a codebook size of $\text{prod}(L) = 1000$.

In summary, our face point grid representation learning model (\autoref{fig:geometry-learning}) compresses each 2D face point grid into $2\times 2$ tokens.
Edge point grids are compressed similarly into $2$ tokens.
As in BrepGen, each face and edge point grid is normalized into a $[-1, 1]^3$ box.
This allows the learned representations to be invariant to translation and scaling of the point grids.
The model is trained with a mean-squared error loss comparing the predicted and ground truth point grids.

\subsection{Invariance to the UV-origin}
\label{sec:invariance_uv_origin}
Geometrically similar surfaces and curves can be parameterized differently.
Specifically, the $(\umin, \vmin) = (0.0, 0.0)$ origin point in the parameter domain can map to any of the four corners of the point grid.
Similarly for edges, the start point of the curve can be at either end.
Choosing a different origin point for the point grid, is equivalent to applying flips and rotations along the u- or v-axis of the point grids.
UV-Net~\cite{jayaraman2021uvnet} first observed this problem and solved it using $\text{D}_2$ equivariant convolutional layers.
We instead propose a simple normalization procedure: 
considering the points in the four corners of the point grids, we sort them using lexicographic sorting of the x, y and z coordinates, and ensure that the corner that is lowest in the sorted order is the UV-origin by flipping the point grids.
This normalization method has the advantage of not needing to change the neural network or introduce augmentations that can slow down training.
\section{Topology Representation}
\label{sec:topology-representation}
Different from the global-coordinate ordering used in prior works \cite{nash2020polygen,jayaraman2022solidgen}, we use a breadth-first traversal of the B-Rep face adjacency graph (\autoref{sec:bft}). Topological relationships are progressively encoded into special reference tokens following the traversal order (\autoref{sec:local-ref}).

\subsection{Breadth-First Traversal} 
\label{sec:bft}
B-Rep topology can be viewed as a face adjacency graph where B-Rep faces are the nodes and B-Rep edges are the connections between faces.
The top row in \autoref{fig:data} illustrates a breadth-first traversal (BFT) over this graph that naturally yields a sequential representation of the B-Rep topology. It begins from the bottom face $F_0$ and proceeds to the two cylindrical side faces $F_1$ and $F_2$. The two sides are connected to the bottom by edge $E_{1,0}$ and $E_{2,0}$. They are also connected to each other by edges $E_{2,1}$ and $E^{*}_{2,1}$. The traversal finally reaches the top face $F_3$. Note that the B-Rep graph could contain cycles and parallel edges. For simplicity, we use a single $F$ and $E$ token, but recall that each face is encoded into 4 tokens, and each edge into 2 tokens. The exact token definition is in ~\autoref{sec:pretrain}.

A key advantage of BFT is that it iteratively traverses the graph by levels. A new level includes the set of unvisited neighboring faces along with the edges connecting them back to the previous level. In unconditional generation, we define the starting level as the bottom-leftmost face in the solid ($F_0$ in ~\autoref{fig:data}). In B-Rep autocompletion, the starting level includes the user-provided faces and edges (see~\autoref{sec:sft}). 
%
If multiple faces exist in the same level, we sort them based on bounding-box coordinates and traverse in ascending order. 
Any edge that connects same-level faces are also added to the sequence (i.e., $E_{2,1}$ and $E^{*}_{2,1}$). After an edge is visited, both of its connected faces are known. In practice, we tokenize the edge immediately after its connected faces and group them together. This adheres to the top-down view that edges come from face intersections. We use special sentinel tokens to denote the start and end of a face or a  BFT level.

\subsection{Local Topological Reference} 
\label{sec:local-ref}
When navigating the B-Rep face adjacency graph, edges frequently reconnect to faces from earlier levels, requiring a method to reference these previously visited faces.
By the definition of breadth-first traversal in ~\autoref{sec:bft}, an edge at level $l$ can only connect to faces from the previous level. This limits the possible face-edge incidence to a local window spanning only the two most recent levels: edges can only connect to faces at level $l{-}1$ or earlier faces within the same level $l$. This motivates us to use a local topological reference $T$ that only encodes the necessary face-edge incidence within the two most recent BFT levels. 

%
Referent tokens are defined as special tags that refer to unique faces within the local context window. Practically, $T$ represents a numerical face ID, with the first face in the local window assigned $T_0$, the next face $T_1$, and so on. As illustrated in the bottom row of ~\autoref{fig:data}, this local window dynamically shifts during breadth-first traversal. Early in the traversal, the local context (green window) encompasses three faces \{$F_0,F_1,F_2$\}, each tagged with references \{$T_0,T_1,T_2$\}. To encode the edge connection for $E_{2,1}$ (an edge between $F_1$ and $F_2$), both faces need to be referenced. $F_2$ is directly inferred as it is the first face before $E_{2,1}$. Since edges are added immediately after all their connected faces have been traversed, the first face before an edge will be one of its connecting faces. Consequently, we only need to encode a topology token with tag $T1$ that points to $F_1$.

As the local window progresses to the next two BFT levels, the reference tokens are reset and updated. In the next local context (blue window) shown in ~\autoref{fig:data}, faces \{$F_1$, $F_2$, $F_3$\} are tagged with references \{$T_0, T_1, T_2$\}. Edge $E_{3,1}$ also connects to $F_1$, but since $F_1$ is now within a new context, the encoded topology token is updated to $T_0$ instead of the previous $T_1$. ~\autoref{sec:ablation-local} demonstrates that using local (as opposed to global) references is crucial for performance.

\begin{figure*}[t]
    \begin{center}
    \includegraphics[width=0.99\textwidth]{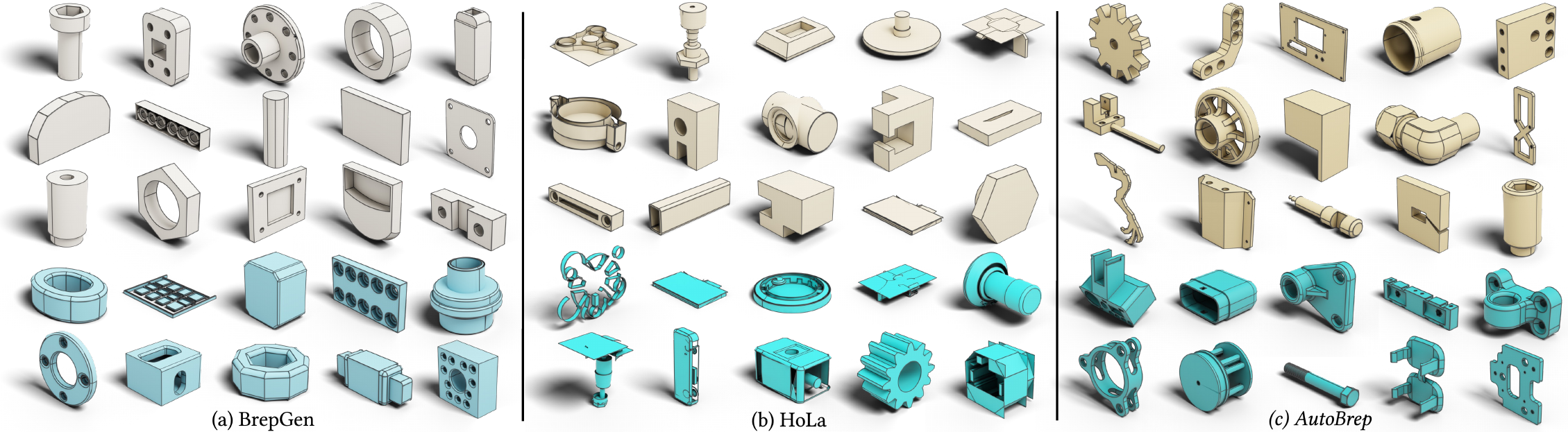}
        \caption{Randomly sampled unconditional results from (a) BrepGen~\cite{xu2024brepgen}, (b) HoLa~\cite{HolaBRep25}, and (c) our method \modelName. Last two rows highlight the generated hard complexity solids (blue color). Our method generates high-quality B-Reps with better complexity, diversity and watertightness. }
        \label{fig:uncond_autobrep}
    \end{center}
\end{figure*}

\section{\modelName: Unified B-Rep Generation}
\label{sec:model}

\subsection{Unconditional Pre-Training}
\label{sec:pretrain}
Breadth-first traversal incrementally constructs a sequence of face, edge, and topology tokens as it iterates the B-Rep face adjacency graph. After converting all B-Reps into this unified tokenized representation, we pretrain a large autoregressive Transformer as the base model via next-token prediction using the standard cross-entropy loss.
The vocabulary used in \modelName\ includes key components representing different aspects of the B-Rep geometry and topology. 

\paragraph{Bounding-Box Coordinate ($C$)} 
As described in \autoref{sec:fsq}, we normalized the face and edge point grids to a canonical space which removes any global positional information. To correctly place the primitives during generation, each geometry is paired with a 3D bounding box that specifies its global location and dimension within the solid. We quantized the $[-1, 1]^3$ into 1024 equally spaced bins along each axis and add the bin indices to vocabulary. The bottom left and top right corners of the quantized bounding-box form the six discrete position tokens $C = [x_0,y_0,z_0, x_1,y_1,z_1]$.

\paragraph{Latent Geometry ($G$)}
Every set of bounding-box tokens is paired with latent geometry tokens as defined in \autoref{sec:fsq}. A face geometry consists of 4 tokens, while an edge geometry has 2 tokens. Geometry vocabulary encompasses the FSQ codebook indices of both faces and edges. The pair of tokens $[C, G]$ uniquely defines a face or edge in the solid. In ~\autoref{fig:data}, we use symbol $F$ to represent the face tokens $[C_f, G_f]$, and $E$ to denote the edge tokens $[C_e, G_e]$. Note that the bounding-box tokens $C$ are generated for every face and edge primitive prior to the latent geometry $G$ . This enables the model to understand each primitive’s position, size, and spatial relationships relative to others, allowing it to effectively associate spatially close but topologically distant primitives in the sequence.

\paragraph{Topology Reference ($T$)}
Local reference tokens from~\autoref{sec:local-ref} consist a set of face tags $[T_0,T_1,T_2...]$ defined within a local context window. We allocate 200 unique face IDs for the reference tags which can support B-Reps with up to 200 faces across two breadth-levels. We explicitly avoid a deterministic sequence order by applying random rotation during training (see ~\autoref{sec:train_details}). This changes the initial face and alters the traversal order for the same solid. We find that training with different traversals reduces overfitting to a particular token sequence. 

\paragraph{Special Tokens}
Six sentinel tokens indicate the start and end for: the sequence, the B-Rep tokens, a BFT level, and a face. We also introduce complexity meta tokens based on face count (\textbf{E}asy, \textbf{M}edium, \textbf{H}ard, \textbf{R}andom), enclosed within \texttt{<META>}, \texttt{<\textbackslash META>} for extra control. Concretely, easy samples have <25 faces, medium samples have 25--50 faces, and hard samples have >50 cases. Finally, a dummy unassigned reference token $T_{u}$ is added for autocomplete (see ~\autoref{sec:sft}). 


\paragraph{Level-based Attention Dropout}
Long B-Rep sequences representing complex solids are prone to overfitting. To improve generation at test time, we apply a level-based attention dropout during pre-training. Concretely, when predicting the tokens at level $l$, an attention mask is randomly applied to drop tokens outside the local context window, ranging from first level to $l{-}2$. This encourages the model to focus on local topological structures and be less dependent on global geometry and topology features.

\subsection{Supervised Fine-tuning for Autocomplete}
\label{sec:sft}
Automatically constructing a solid which connects some user specified faces to build a watertight B-Rep body is a useful task introduced in \cite{xu2024brepgen}.  When parts are designed to fulfill a role in a larger assembly, designers often require the generated solids to attach to existing geometry.
After pretraining the base model, we perform supervised fine-tuning for controllable B-Rep autocompletion. Given a partial B-Rep from the user, \modelName\ conditionally generates the missing topology and geometry to form a valid solid. We tokenize the user input following ~\autoref{sec:geom-representation}, ~\autoref{sec:topology-representation} and ~\autoref{sec:pretrain}. User tokens are appended to the start as part of the first BFT level. During training, random faces on the solid are also selected as user tokens in the first level (see~\autoref{sec:train_details}). Similar to pre-training, this leads to different traversals and avoids a fixed sequence order. Loss is only applied on the generated levels. Since we treat user input as another level in our unified representation, we do not need to regenerate these faces, thus guaranteeing their exact preservation in the valid results. 

A minor modification is required to handle dangling edges. During breadth-first traversal, some edges in the user level will not be connected to any face, since they are to be generated in later levels. We use a special dummy token $T_{u}$ to indicate an edge has unassigned face reference. The model generates the unassigned connections in the second level with the correct reference tokens. 
Since \modelName~works with the $[-1,1]^3$ normalized space, during test time, we additionally require a bounding box defining the domain for generating autocompletions. To address the alignment issue between training and real-world user-provided boxes, we randomly augment training data with up to 15\% translation and scaling.

\section{Dataset}
\label{sec:abc-data}
Two new datasets are constructed based on ABC~\cite{koch2019:abc}: ABC-1M and ABC-Constraint. We use them to pretrain the base model and finetune for autocomplete, respectively.

\subsection{ABC-1M}

ABC-1M is extended from the original ABC dataset \cite{koch2019:abc}, which includes both single-body and multi-body CAD models. Multi-body files were decomposed into individual parts, resulting in \textasciitilde3.1 million single-part STEP files. To remove duplicates, we grouped parts by a hash based on part name, edge/vertex counts, and primitive types (cones, cylinders, etc.). Within each group, parts were considered duplicates if their surface areas and volumes (scaled by bounding-sphere) were within 1\% relative to the maximum values. This leads to \textasciitilde1.3 million unique solids. ABC-1M is divided into train/val/test sets (70\%/15\%/15\%) using a stratified sampling based on face count to ensure balanced geometric complexity across splits.
During preprocessing, following prior work~\cite{xu2024brepgen}, we split all periodic surfaces along their seam edges. This for example, converts a cylinder into two half cylinders and simplifies post-processing required to fit parametric surfaces.

\subsection{ABC-Constraint}
To train \modelName~for B-Rep autocompletion as described in \autoref{sec:sft}, we need a way to identify likely faces that users might select as input.
Assembly interfaces are typically the faces designers want to specify for the autocomplete task.  A common way parts are connected is by holding planar faces together using bolts.   Consequently, for mechanical parts like those found in the ABC dataset, cylindrical bolt holes and planar faces on the convex hull of a solid are more likely to be the faces specified by designers.  To this end, we identify cylindrical bolt holes as concave cylinders with their axis aligned with the normal of an adjacent planar face.  Planar faces which are coincident with the axis aligned bounding box of the solid are also identified. We detect all likely user constraints from ABC-1M and store them as ABC-Constraint for training B-Rep autocompletion. 


\begin{table}
\caption{Quantitative evaluations for unconditional generation. HoLa and BrepGen are retrained on ABC-1M, while DTGBrepGen* uses their published model. We report metrics for \textit{Coverage} (COV) percentage, \textit{Minimum Matching Distance} (MMD), \textit{Jensen-Shannon Divergence} (JSD), \textit{Novel}, \textit{Unique} and \textit{Valid}. MMD and JSD scores are multiplied by $10^2$. Inference time is averaged per-face for valid B-Reps. \textit{AutoBrep coord} ordering follows the sorted face bounding-box coordinates. \textit{AutoBrep global} removes local context window and replaces local topological reference with a global reference. }
\label{tab:abc1m_uncond}
\begin{center}
\setlength\tabcolsep{3.5 pt}
\small
\begin{tabular}{lccc|ccc|c}
\toprule
       Method  & COV   & MMD  & JSD  & Novel   & Unique & Valid  & Time   \\
         &  \% $\uparrow$  & $\downarrow$  & $\downarrow$  &\% $\uparrow$   & \% $\uparrow$  & \% $\uparrow$  & $\downarrow$ \\ \midrule 
         
DTGBrepGen*  &  59.39      &  1.60  &   1.47   &  - & - & 64.3 & 0.69 \\
BrepGen    &  67.41      &  1.91    &   3.50   & 99.7 & 94.5 & 46.6 & 1.25\\
HoLa  &  67.80      &  1.62    &   2.86   & 99.8 & 95.3 & 54.8 & 1.68 \\
\midrule  
\textit{AutoBrep coord}  & 66.28      &  1.53   &  1.61   & 99.9 & 95.8  & 63.6  & 0.49 \\
\textit{AutoBrep global}  & 68.09      & 1.58     &  1.03   & 99.7 & 94.5  & 65.3 & 0.48\\
\modelName\ & \textbf{71.49}  & \textbf{1.45}  & \textbf{0.97}   & 99.8 & 93.7 & \textbf{70.8} & \textbf{0.46}\\
\bottomrule
\end{tabular}

\end{center}
\end{table}

\section{Experiments}
\label{sec:experiment}

This section presents unconditional B-Rep generation and user-controlled autocompletion results. Extensive evaluations show that \modelName\ outperforms baselines with better quality and valid ratio.

\subsection{Implementation Details}

\subsubsection{Network Architecture}
Our base model is a GPT-style Transformer \cite{vaswani2017transformer} with 16 self-attention layers, 32 attention heads, 2048 hidden dimension, and 3K context length. Swish GLU~\cite{shazeer2020glu}, RMS layer normalization~\cite{zhang2019root}, rotary positional embeddings~\cite{su2024roformer}, flash attention~\cite{dao2022flashattention}, and grouped query attention~\cite{ainslie2023gqa} with 8 key-value heads are used for faster inference. Dropout of $0.1$ is applied to feedforward and embedding layers.

\subsubsection{Training}
\label{sec:train_details}
We pre-trained on ABC-1M data with maximum 100 faces and 1,000 edges, excluding non-manifold solids and those with tiny faces or edges below quantization precision. Sequences longer than 3,000 tokens are randomly truncated on the left or right. The complexity meta token is dropped 20\% of the time to support unconditional generation. Data is augmented with 90, 180, or 270 degree rotation around one axis. Four level FSQ codebook of $[8,5,5,5]$ is used for the face and edge autoencoders. Level-based attention dropout is set to 0.1. We trained on 64x Nvidia H100 GPUs with a batch size of 256 and mixed precision. Model is optimized with AdamW \cite{loshchilov2018decoupled} with a learning rate of $3e{-}4$ and $0.05$ weight decay. Learning rate is lowered to $5e{-}5$ after 100 epochs and continue for 50 more epochs. Easy data is dropped 90\% of the time in the last stage of training. 

For autocompletion, the base model is further fine-tuned on ABC-Constraint. To maintain generation quality, we split training to 80\% autocomplete, 20\% unconditional generation. To avoid overfitting to detected constraints, a random subset of faces is sampled per solid as the user input. We use a mixture of 70\% constraints to 30\% random faces.

\subsubsection{Inference}
We use top-p sampling~\cite{holtzman2020toppsampling} of $0.9$ and temperature of $1.0$ for sampling. Half-precision and key-value caching are used to improve speed. Post-processing model output to B-Rep follows BrepGen~\cite{xu2024brepgen} where UV grids are approximated by B-spline parametric surfaces and curves, and then sewn back into watertight solids. Face-edge adjacency is directly decoded from the generated topological reference tokens.

\subsubsection{Evaluation Metrics}
Following prior works~\cite{xu2024brepgen,HolaBRep25,wu2025cmtcascademartopology}, we use two types of metrics to quantitatively measure the generation quality: \textit{Point-based Metrics} and \textit{CAD-based Metrics}. Point-based  metrics uniformly sampled 2,000 points from the surface and compute the metrics for \textit{Coverage} (COV), \textit{Minimum Matching Distance} (MMD), and \textit{Jensen-Shannon Divergence} (JSD). CAD-based metrics hashed the generated B-Reps and report the percentage of \textit{Novel} and \textit{Unique} data. An averaged \textit{Valid} ratio is also used to measure the percentage of \textit{watertight} B-Reps. 

\begin{figure}
    \centering
    \includegraphics[width=0.98\linewidth]{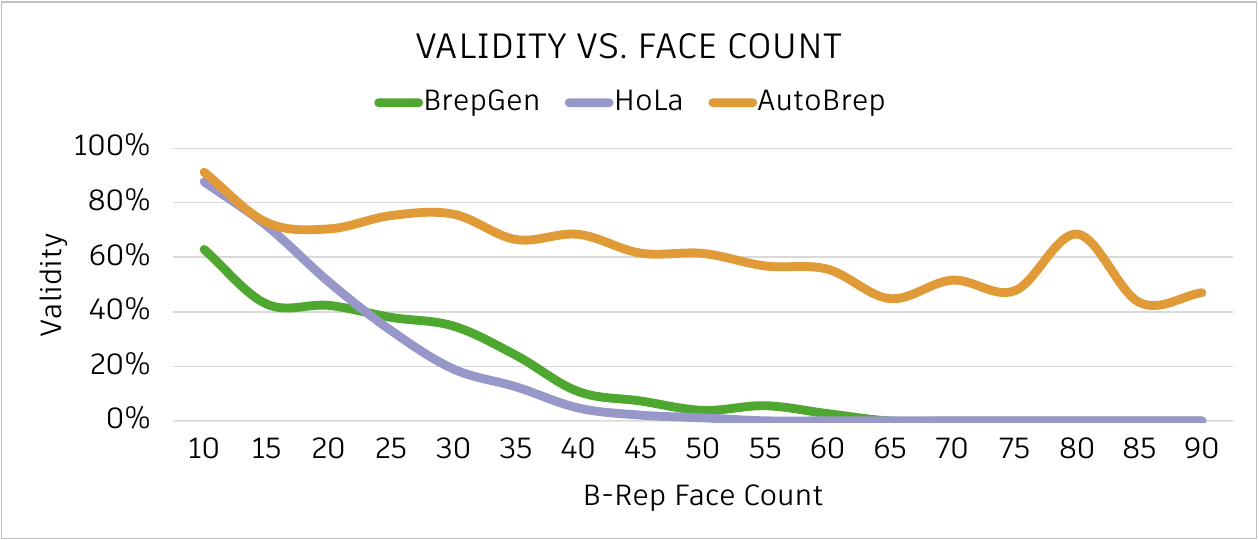}
    \caption{Validity of generated B-Reps as a function of the face count.}
    \label{fig:validity_vs_faces}
\end{figure}

\begin{table}
\caption{Codebook usage and per-point Root Mean Square Error (RMSE) on the ABC-1M test set. RSME value is multiplied by $10$. VQ-restart refers to VQ-VAE variant with random restart.}
\label{tab:codebook}
\begin{center}
\setlength\tabcolsep{3.5 pt}
\small
\begin{tabular}{@{}lcccc@{}}
\toprule
          & \multicolumn{2}{c|}{Face Codebook}                    & \multicolumn{2}{c}{Edge Codebook}                    \\
          & \multicolumn{1}{c}{RMSE $\downarrow$} & \multicolumn{1}{c|}{Usage $\uparrow$} & \multicolumn{1}{c}{RMSE $\downarrow$} & \multicolumn{1}{c}{Usage $\uparrow$} \\ \midrule
VQ         &  0.473 $\pm$ 0.469  & 99.9\%   &  0.339 $\pm$ 0.108   & 98.5\%  \\
VQ-restart &  0.338 $\pm$ 0.402   & 100\%   & 0.286 $\pm$ 0.094    & 99.3\%  \\
\textit{FSQ}        & \textbf{0.075 $\pm$ 0.108}    & \textbf{100\%}   & \textbf{0.154 $\pm$ 0.065}   & \textbf{100\%}  \\  \bottomrule

\end{tabular}

\end{center}
\end{table}

\subsection{Unconditional Generation}
For fair comparison, we retrained BrepGen~\cite{xu2024brepgen} and HoLa~\cite{HolaBRep25} on ABC-1M with a maximum of 100 faces per solid, and report the qualitative and quantitative results.

\subsubsection{Qualitative Evaluation}
\autoref{fig:uncond_autobrep} shows randomly sampled results for unconditional B-Rep generation, with hard complexity solids highlighted in the last two rows.
\modelName\ is sampled with the complexity meta token set to \textit{random}. Results demonstrate that our model generates higher-quality B-Reps compared to the baselines. Geometry is more diverse, featuring chamfers and fillets. Topology is also more complex and resembles real-world CAD models. We observe the most notable improvements on hard complexity solids, where the generated results are more realistic than BrepGen, and exhibits fewer open surfaces or self-intersections than HoLa.

\subsubsection{Quantitative Evaluation}
We report point-based metrics using ABC-1M test set as the reference. For efficiency, novel percentage is averaged over 50,000 randomly sampled training data. We fail to retrain DTGBrepGen~\cite{DTGBrepGen} on ABC-1M due to convergence issues and resource constraints. Results are reported with their pretrained checkpoint on a ABC subset of 30 maximum faces. CAD-based metrics is also not reported as it uses a B-spline representation. First three columns in \autoref{tab:abc1m_uncond} shows the point-based metrics. \modelName\ outperforms baselines with better coverage and lower MMD, JSD scores, indicating results are more aligned with ground-truth data. For novel and unique, we follow prior work~\cite{xu2024brepgen} and identify similar B-Reps after 4 bit quantization. Results show that our model has high novel and unique scores, demonstrating that the generation is diverse and not overfitting. 

Our method also achieves the fastest inference speed, computed as the averaged time in seconds to generate a face for valid B-Reps (see last column in ~\autoref{tab:abc1m_uncond}). We test it on a \textit{g5.12xlarge} instance on AWS with Nvidia A10G GPU and include the post-processing and rebuild time. Consistent with our training, DTGBrepGen, HoLa and BrepGen baselines are also trained with mixed precision and sampled with half-precision. All three baselines are diffusion-based and do not benefit from the use of KV caching.

\modelName\ has the highest averaged valid ratio of 70.8\%. A B-Rep is considered valid if it passes the solid modeling kernel's checks for correct geometry and watertight topology. To better understand the improvement in watertightness, we plot the valid ratio against the face count in \autoref{fig:validity_vs_faces}. We see that baseline methods struggle with validity as face count increases, while our model remains highly scalable and maintains \textasciitilde$50\%$ valid ratio for solids up to 100 faces.

\begin{table}
\caption{Perplexity (PPL) score under different meta complexity. Results are evaluated on the ABC-1M test set. Easy is B-Rep with less than 25 faces, Medium is 25 to 50 faces, Hard is 50 to 100 faces.}
\label{tab:token}
\begin{center}
\setlength\tabcolsep{3.5 pt}
\small

\begin{tabular}{@{}lcccc@{}}
\toprule
PPL  $\downarrow$           & \multicolumn{1}{c}{Random} & \multicolumn{1}{c}{Easy} & \multicolumn{1}{c}{Medium} & \multicolumn{1}{c}{Hard} \\ \midrule
\modelName\ coord  & 3.29  &  2.75    & 3.35   & 3.31  \\
\modelName\ global & 3.15  & \textbf{2.56} & 3.07 &  2.96  \\
\modelName\    & \textbf{2.98}   &  2.80   &  \textbf{2.94}  & \textbf{2.88}     \\ \bottomrule
\end{tabular}

\end{center}
\end{table}

\begin{figure}
  \centering
  \includegraphics[width=0.98\linewidth]{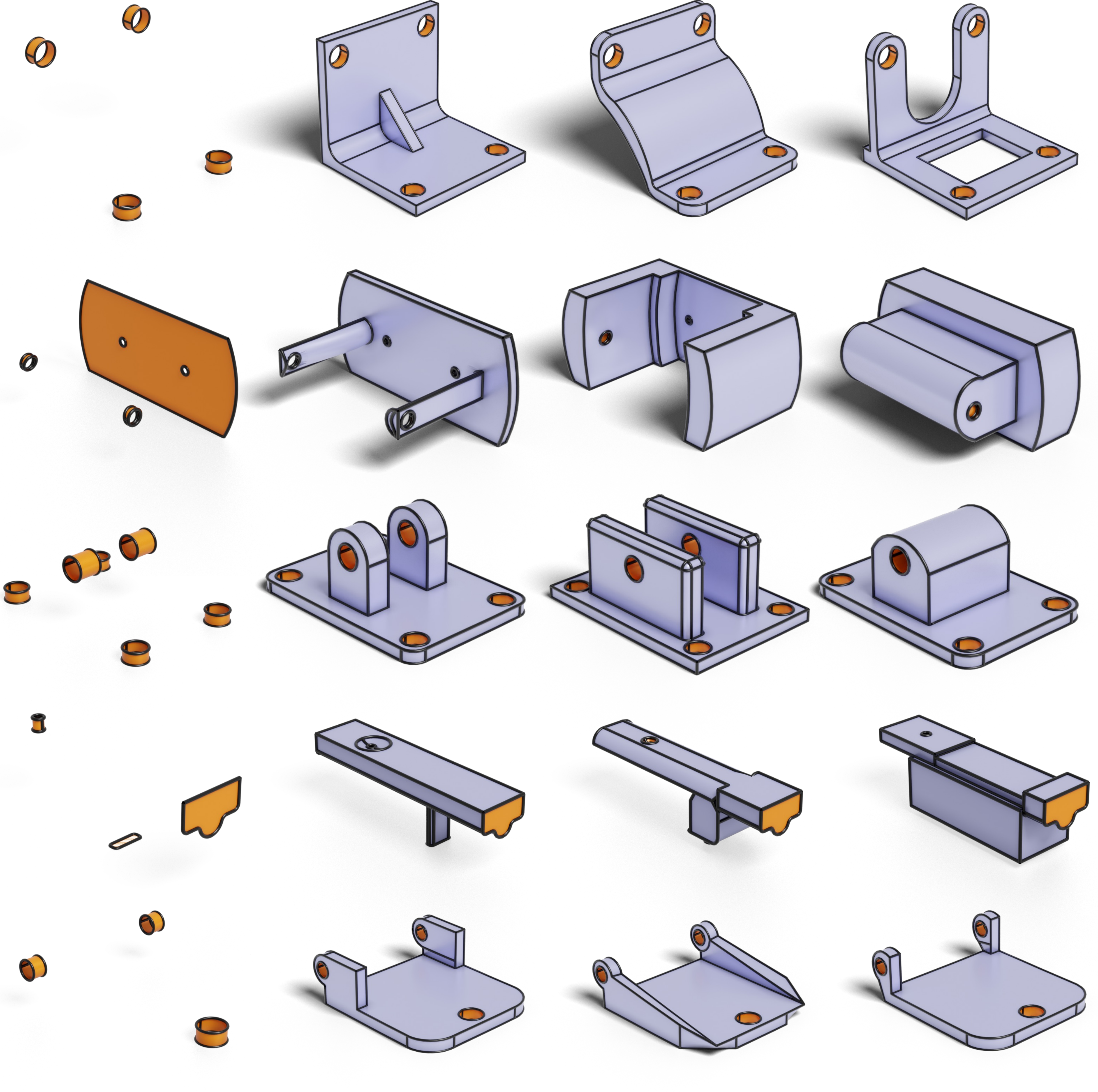}
  \caption{Autocompletions from faces representing the assembly interfaces (orange), showing that a wide variety of shapes and styles can be generated.}
  \label{fig:autocomplete}
\end{figure}

\subsubsection{Ablation Study on Topology and Traversal Order}
\label{sec:ablation-local}
We remove the local context window in ~\autoref{sec:local-ref} and retrain \modelName\ with global topology. This is done by not resetting the reference tags based on BFT levels and increment the references by counting from the first face till the end. Results in \autoref{tab:abc1m_uncond} (\textit{AutoBrep global}) show that global topology reference reduce performance on almost all metrics.
This highlights the benefit of learning local structure for improving generation quality and validity. We also train a variant where the sequence order follows the sorted face bounding-box coordinates. Results in \autoref{tab:abc1m_uncond} (\textit{AutoBrep coord}) show that sorting B-Rep faces by their global coordinate positions significantly degrades performance, resulting in a lower validity ratio and reduced coverage. The substantially higher JSD further indicates that the generated distribution significantly deviates from ground-truth.

\subsubsection{Ablation Study on Codebook}
We compare our FSQ codebook with VQ-VAE~\cite{vqvae} and its variant using random codebook restart~\cite{dhariwal2020jukebox}. For fair comparison, the two VQ baselines also use Deep Compression AutoEncoder as the backbone. Latent dimension and total codebook size are kept the same as those in \modelName. ~\autoref{tab:codebook} reports the codebook usage and RMSE on the ABC-1M test set. Results show that FSQ achieves the lowest RMSE per-point for both the face and edge encoding, while having the highest codebook usage of 100\%.

\subsubsection{Token-Centric Evaluation}
\autoref{tab:token} presents the perplexity metrics for \modelName\ and its two variants based on global topology and bounding-box coordinate ordering. The results show that the default \modelName\ achieves the lowest perplexity on Random, Medium, and Hard complexity data, whereas the coordinate ordering variant yields the highest perplexity on almost all levels. This indicates that the default \modelName\ is more effective at modeling complex solids, whereas the two variants are largely limited to simpler shapes.

\subsection{B-Rep Autocompletion}
\label{sec:exp-autocomplete}
Figure \ref{fig:autocomplete} shows some examples of autocompletions generated by the model for common patterns of bolt holes found in mechanical parts.
Our model effectively learns to generate the correct geometry and the surrounding topology that connects them to the user provided faces.
Results show the wide variety of useful shapes the model can generate, exactly matching the specified assembly interfaces (something that is not possible with prior work) while offering designers a range of choices in terms of material usage and style.   

\begin{figure}[t]
    \centering
    \includegraphics[width=0.98\columnwidth]{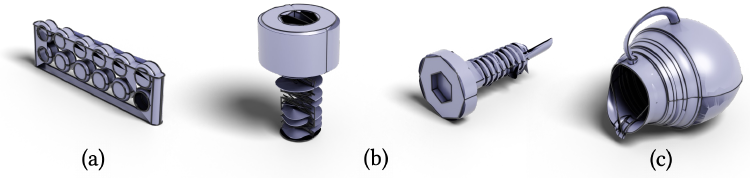}
    \caption{Failure cases for \modelName~ include (a) self-intersections, (b) screws with incorrectly generated threads , and (c) thin shells with artifacts.}
    \label{fig:fail}
\end{figure}

\subsection{Failure Cases} 
\autoref{fig:fail} shows B-reps generated by \modelName\ from three common failure case categories. 
Self intersections (\autoref{fig:fail} (a)) can occur when lots of B-Rep faces are packed in narrow volumes, but the B-Rep can often be rebuilt from the model output even in these cases despite lack of watertightness. The sewing operation in OpenCascade is relatively robust to these minor inconsistencies. Threads in screws (\autoref{fig:fail} (b)) have faces that are long and winding in one direction and narrow in the other, making it challenging to represent with point grids.
Similarly, thin shells (\autoref{fig:fail} (c)) are challenging owing to limited precision in the quantization, and the presence of sliver faces.


\section{Conclusion}
We have presented \modelName, a simple and highly scalable model that enables high quality, complex B-Rep generation.
Our method works by tokenizing faces, edges and their topology into a unified set of discrete tokens that can be learned by a single autoregressive generative model.
This approach simplifies training, enhances inference speed, reduces cumulative errors, and is easily scalable, unlike previous methods that relied on multiple neural networks and multi-stage training.
Our model natively supports B-Rep autocompletion with exact preservation of input geometry, an important feature for generating parts in the context of an assembly.

In future work, we plan to explore strategies to address failure cases. One straightforward approach is to filter out B-Reps from the training data that are overly complex and not worthwhile to generate. For instance, threads are typically part of standard components like bolts with well-defined specifications that can be sourced from catalogs rather than generated. Thin shells can be better managed by enhancing the precision with which bounding boxes are delineated. In the long term, we are keen to investigate alternative geometry representations beyond point grids that may be better suited for mechanical components.

\bibliographystyle{ACM-Reference-Format}
\bibliography{bibliography}

\appendix

\begin{figure*}
    \centering
    \includegraphics[width=0.98\textwidth]{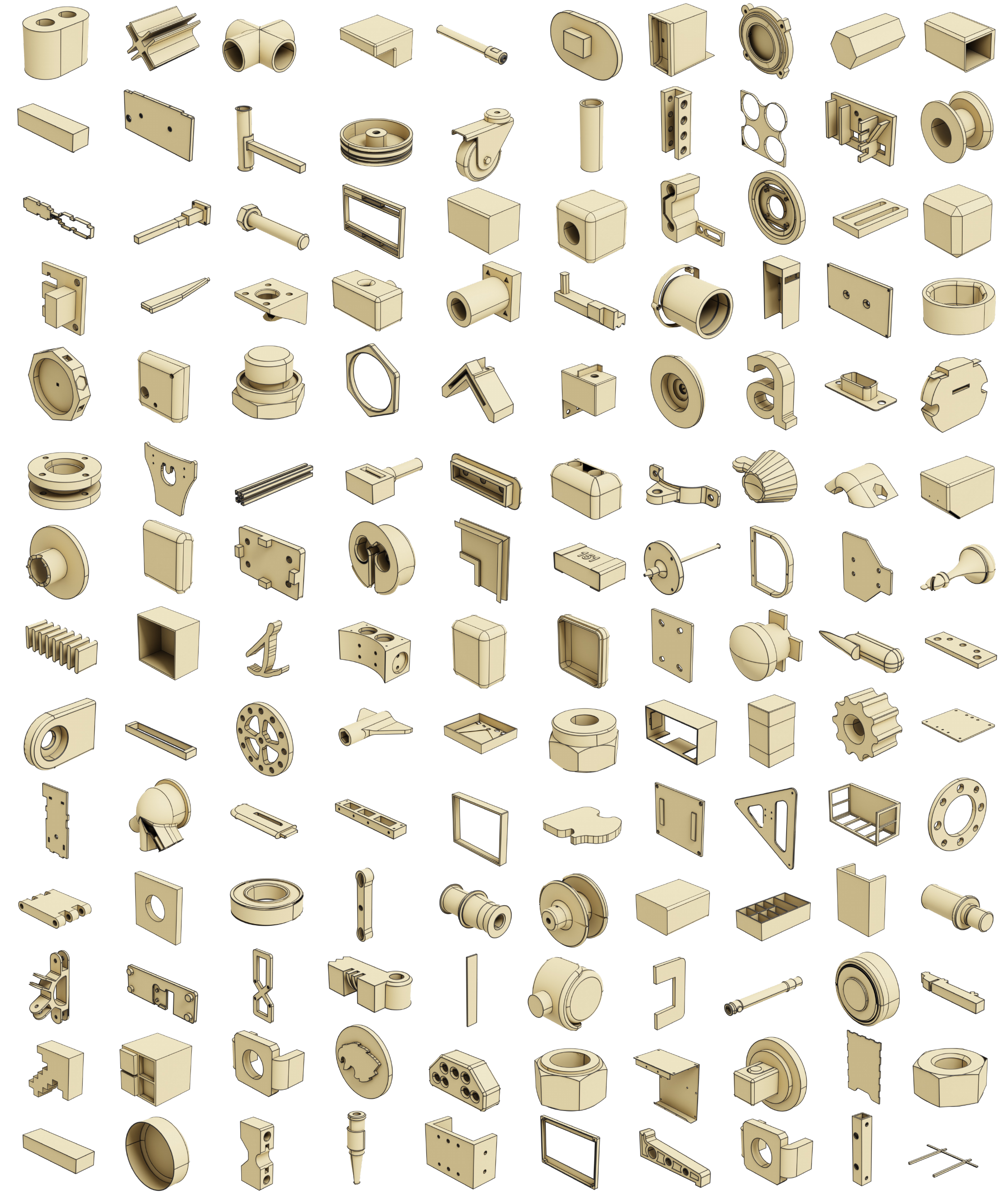}
    \caption{B-Reps sampled unconditionally from \modelName~trained on the ABC-1M dataset.}
    \label{fig:uncondmosaic}
\end{figure*}

\begin{figure*}
    \centering
    \includegraphics[width=0.98\textwidth]{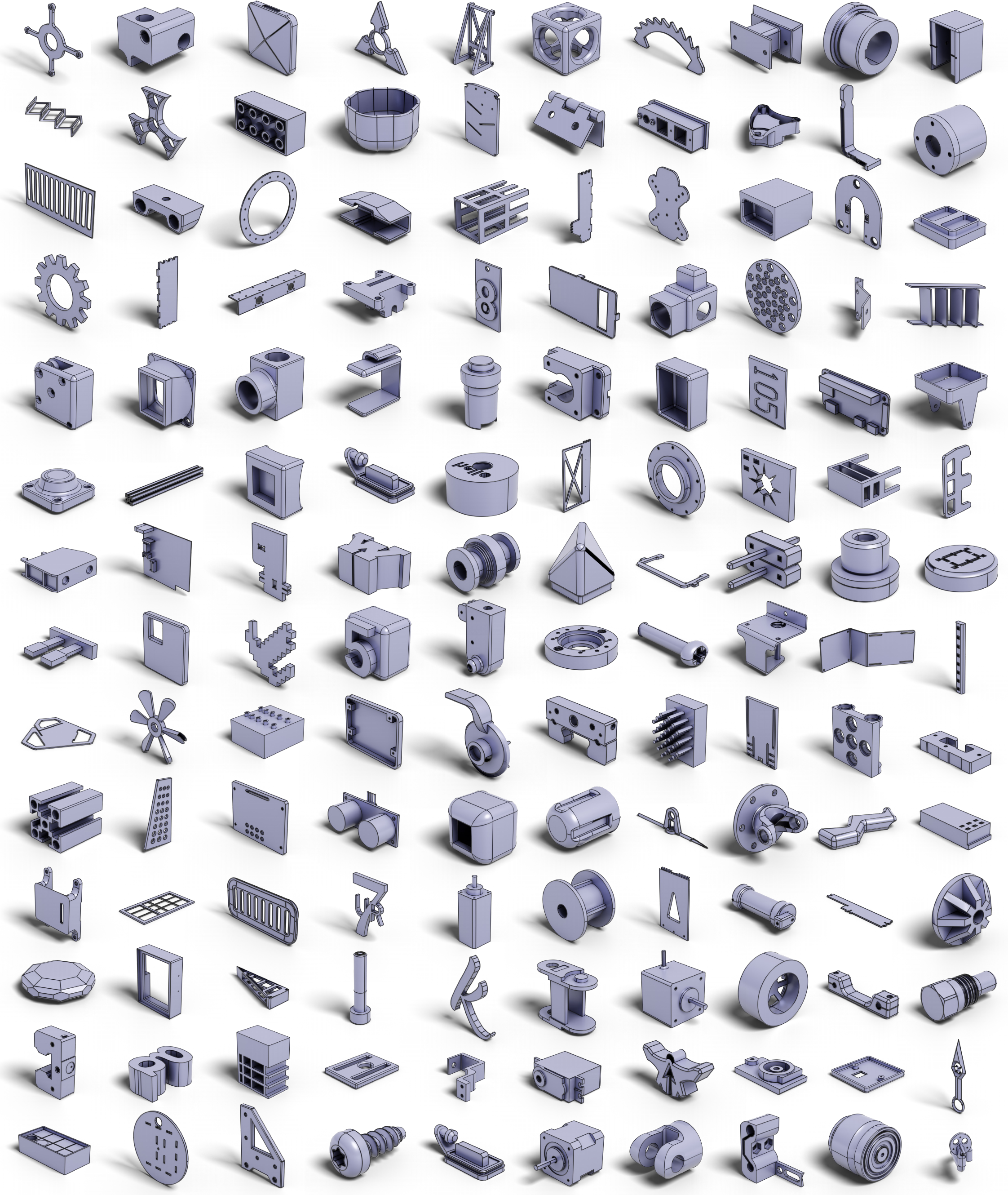}
    \caption{B-Reps sampled with Hard complexity token from \modelName~trained on the ABC-1M dataset.}
    \label{fig:hardcondmosaic}
\end{figure*}



\end{document}